\documentclass[lettersize,journal]{IEEEtran}
\usepackage{amsmath,amsfonts}
\usepackage{algorithmic}
\usepackage{algorithm}
\usepackage{array}
\usepackage[caption=false,font=normalsize,labelfont=sf,textfont=sf]{subfig}
\usepackage{textcomp}
\usepackage{stfloats}
\usepackage{url}
\usepackage{verbatim}
\usepackage{graphicx}
\usepackage{cite}
\usepackage{amssymb}

\usepackage{amsfonts}

\usepackage{enumerate}
\usepackage{multirow}
\usepackage{threeparttable}
\usepackage{makecell}
\usepackage{color}
\usepackage{bm}

\usepackage{booktabs}
\usepackage[colorlinks,
linkcolor=blue,
anchorcolor=blue,
citecolor=blue]{hyperref}

\DeclareRobustCommand*{\IEEEauthorrefmark}[1]{%
	\raisebox{0pt}[0pt][0pt]{\textsuperscript{\footnotesize\ensuremath{#1}}}}

\pdfoutput=1

\begin{document}

\title{Online progressive instance-balanced sampling for weakly supervised object detection}
\author{Minghao Chen,Yunong Tian,Zhishuo Li,En Li,Zize Liang}

\author{
	\IEEEauthorblockN{
		Minghao Chen\IEEEauthorrefmark{1}\IEEEauthorrefmark{2},
		Yunong Tian\IEEEauthorrefmark{1}\IEEEauthorrefmark{2},
		Zhishuo Li\IEEEauthorrefmark{1}\IEEEauthorrefmark{2},
		En Li\IEEEauthorrefmark{1}\IEEEauthorrefmark{2}\IEEEauthorrefmark{3} and
	    Zize Liang\IEEEauthorrefmark{1}\IEEEauthorrefmark{2}}
    
	\IEEEauthorblockA{\IEEEauthorrefmark{1}The State Key Laboratory of Management and Control for Complex Systems, Institute of Automation, Chinese Academy of Sciences,  95 Zhongguancun East Road, Beijing 100190, China}
	
	\IEEEauthorblockA{\IEEEauthorrefmark{2}The School of Artificial Intelligence, University of Chinese Academy of Sciences, No.19(A) Yuquan Road, Beijing 100049, China}
	
	\IEEEauthorblockA{\IEEEauthorrefmark{3}Engineering Laboratory of Industrial Vision and Intelligent Equipment Technology, Chinese Academy of Sciences, 95 Zhongguancun East Road, Beijing 100190, China }
		
	
	\IEEEauthorblockA{Corresponding Author: En Li \quad Email: en.li@ia.ac.cn}}

\markboth{Journal of \LaTeX\ Class Files,~Vol.~14, No.~8, June~2022}%
{Shell \MakeLowercase{\textit{et al.}}: A Sample Article Using IEEEtran.cls for IEEE Journals}


\maketitle

\begin{abstract}
Based on multiple instance detection networks (MIDN), plenty of works have contributed tremendous efforts to weakly supervised object detection (WSOD). However, most methods neglect the fact that the overwhelming negative instances exist in each image during the training phase, which would mislead the training and make the network fall into local minima. To tackle this problem, an online progressive instance-balanced sampling (OPIS) algorithm based on hard sampling and soft sampling is proposed in this paper. The algorithm includes two modules: a progressive instance balance (PIB) module and a progressive instance reweighting (PIR) module. The PIB module combining random sampling and IoU-balanced sampling progressively mines hard negative instances while balancing positive instances and negative instances. The PIR module further utilizes classifier scores and IoUs of adjacent refinements to reweight the weights of positive instances for making the network focus on positive instances.  Extensive experimental results on the PASCAL VOC 2007 and 2012 datasets demonstrate the proposed method can significantly improve the baseline, which is also comparable to many existing state-of-the-art results. In addition, compared to the baseline, the proposed method requires no extra network parameters and the supplementary training overheads are small, which could be easily integrated into other methods based on the instance classifier refinement paradigm.
\end{abstract}


\section{Introduction}
\label{Introduction}

As a result of the development of convolutional neural networks (CNN), fully supervised object detection (FSOD) \cite{FastRCNN,FasterRCNN,SSDnet,MASKRCNN,Focal_loss} has made great achievements with the assistance of tremendous instance-level datasets. Nevertheless, these datasets were produced with a major expenditure of time and effort. For alleviating this issue, some works \cite{WSDDN,ContextLocnet,MFMIL,WCCN} attempted to adopt datasets with image-level annotations to train networks, which is regarded as the weakly supervised object detection (WSOD) problem. Furthermore, the results of WSOD were utilized as pseudo ground truth bounding boxes \cite{W2F,GradingNet,MPGTBB} to train FSOD networks.

\begin{figure}[h]
	\centering
	\includegraphics[width=3in]{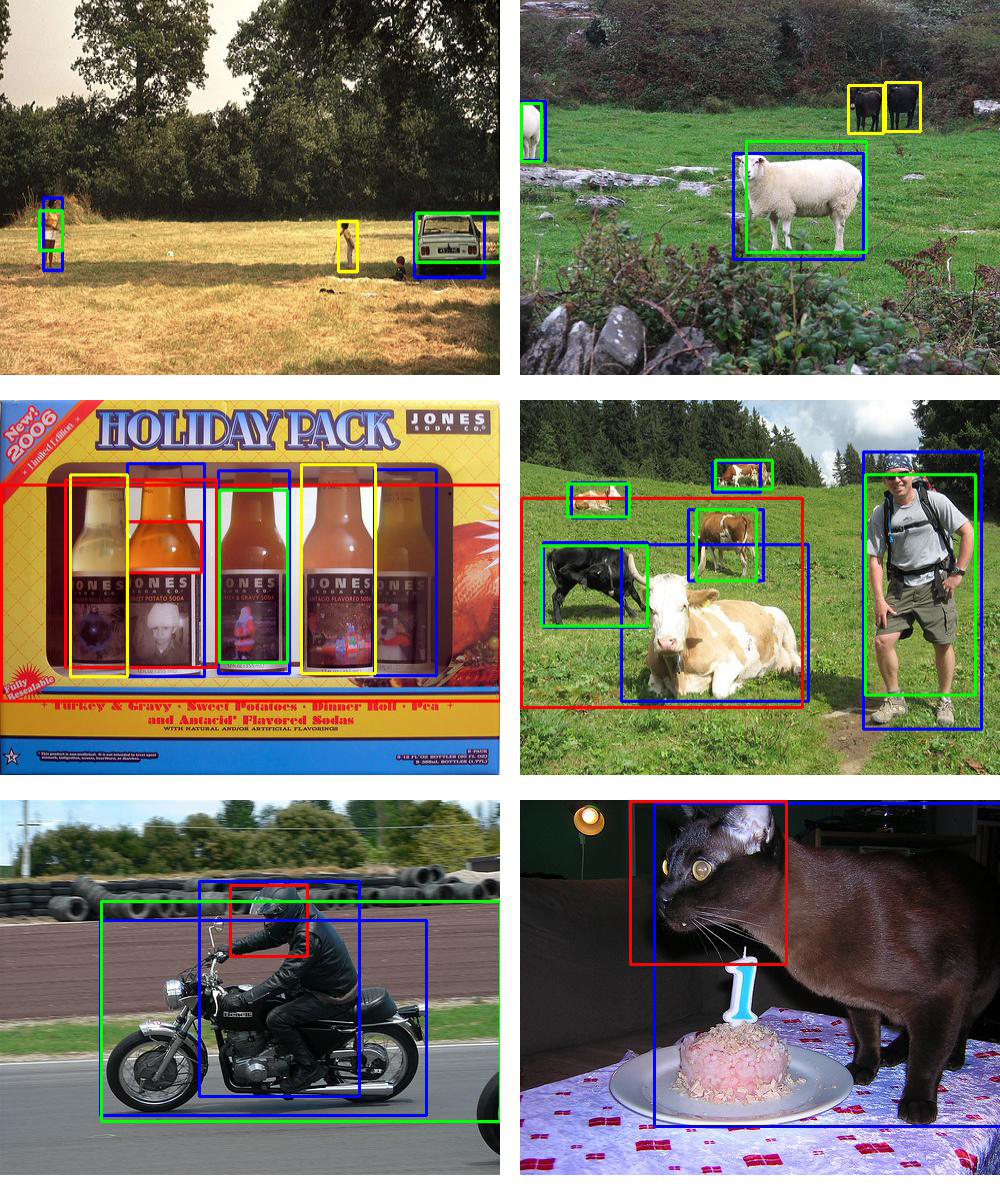}
	\caption{Some detection problems on WSOD. Green rectangles represent the successful detections (IoU$\geq$0.5), red rectangles represent the failed detections (IoU$<$0.5), blue rectangles are the ground truths with detection results, and yellow rectangles indicate the ground truths with no detection results.}
	\label{fig1}
\end{figure}

Nowadays, weakly supervised object detection is considered a multiple instance learning (MIL) problem in many extant methods \cite{WSDDN,OICR,PCL,MELM}. MIL views each image and  object proposal as a bag and an instance respectively. However, as shown in Fig. \ref{fig1}, some challenges \cite{Wetectron,WSOD2021,WSOD2022}, such as location inaccuracy, and instance neglect, still exist when using MIL-based methods in weakly supervised object detection, i.e. the model falls into local minima. 

To alleviate the local minima problem, OICR \cite{OICR} adopted an online instance classifier refinement algorithm to correctly localized objects by spatially overlapped proposals. And some researchers further propose better methods \cite{PCL,BoostOICR,C-MIL,CASD} following this algorithm. However, the fact that the ratio of positive instances to negative instances reaches about 1:100 \cite{OPG} in WSOD while the ratio is held nearly at 1:3 in FSOD is ignored in these methods.

These considerable negative instances will bring up two disadvantages in the training process of OICR as follows: (1) The model is misled by negative instances during training. The ratio of positive instances to negative instances is too small. And the negative instances provide the network with the main gradient contribution. As a result, the network would pay more attention to negative instances, which may cause the model to select the most discriminative object parts as detection results. (2) True positive instances are probably regarded as negative instances. When more than one true positive instance of the same class is in an image, OICR would consider other instances whose IoU between the highest-scoring instance and themselves are less than 0.5 as negative instances. WSRPN \cite{RegionProposal} utilized a simple means that the weight of instances whose IoUs were less than 0.1 was set to zero, which could effectively avoid the happening of the second problem and alleviate the first problem. After that, the ratio would drop to between 1:20 and 1:40. However, it still limits the further improvement of the OICR performance because of the first problem. 

Therefore, the sampling methods are utilized , which are usually used in FSOD to sample examples, to further alleviate the first problem. Random sampling is a straightforward method in FSOD to reduce the number of negative instances, but it is difficult to mine better hard instances. And loss-based sampling methods  such as online hard example mining (OHEM) \cite{OHEM}, were also employed in FSOD to sample hard examples. However, ground truths are lacking in WSOD during the training phase. So the loss-based sampling methods are not befitting WSOD. IoU-balanced sampling \cite{LIBRARCNN} was a simple sampling method only based IoUs of examples, which inspires our work to further mine hard negative instances. Furthermore, positive instances in one class have the same weight during the training phase of OICR, which is unbefitting to select better positive instances. As a consequence, a reweighting method is used to adjust positive instances and balance the positive instances and negative instances from the aspect of soft sampling.

In this paper, an online progressive instance-balanced sampling (OPIS) algorithm is proposed, which is based on the instance classifier refinement paradigm to alleviate the influence of overwhelming negative instances. OPIS algorithm consists of two designs: A progressive instance balance (PIB) module and a progressive instance reweighting module (PIR). The PIB module, which combines random sampling and IoU-balanced sampling, progressively balances positive instances and negative instances. With the ratio of positive instances to negative instances decreasing, IoU-balanced sampling gradually becomes dominant in the PIB module, which helps to mine hard negative instances. Meanwhile, some specific positive instances will be neglected when the corresponding highest-scoring cluster center has no negative instances. The PIR module adopts scores and IoUs of adjacent instance classifiers to reweight the weights of positive instances, which is also helpful for improving the focus of the network on positive instances.

In summary, the key contributions of this work are listed as follows: (1)  An online progressive instance-balanced sampling algorithm based on sampling is proposed. By combining hard sampling and soft sampling, the OPIS algorithm could progressively balance positive instances and negative instances, which alleviates the impact of the overwhelming negative instance problem. (2) The OPIS method is generic. It could be easily applied in other methods based on the instance classifier refinement paradigm. In addition, compared to the baseline, the proposed method does not need an extra network architecture and the appendant training overheads could be negligible, which also allows the model to train in an end-to-end manner. (3) Extensive experiments on PASCAL VOC 2007 and 2012 are executed. The experimental results show the efficacy of the proposed method and the method markedly improves the performance of the baseline, which is comparable to many existing state-of-the-art results.

\begin{figure*}[h]
	\includegraphics[width=7in]{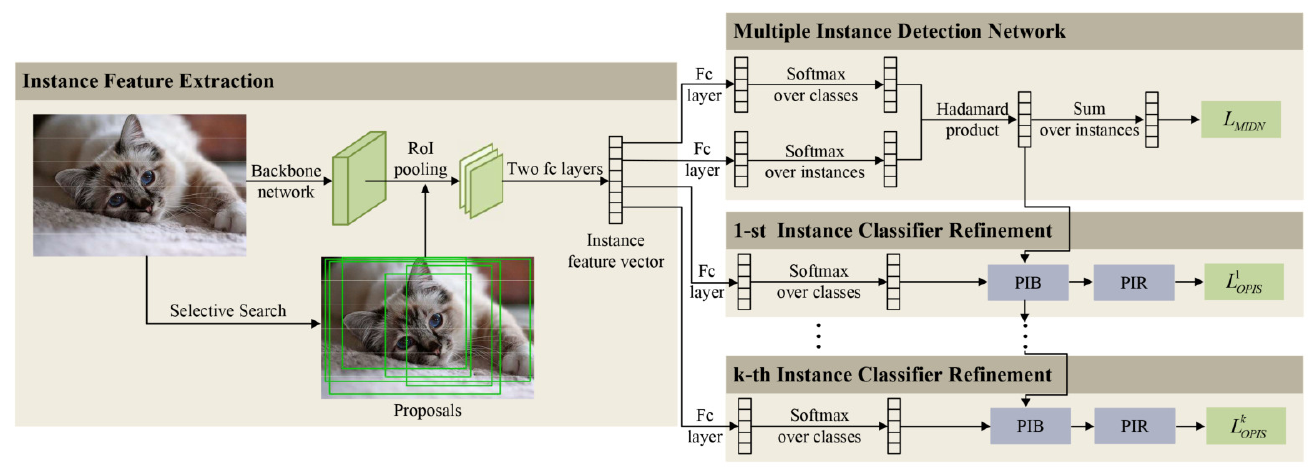}
	\caption{The overall pipeline of the proposed method. It mainly includes three modules, which are instance feature extraction, multiple instance detection network, and K times instance classifier refinements.  }
	\label{fig2}
\end{figure*}

\section{Related work}
\label{Related work}

\subsection{Weakly supervised object detection}

Multiple instance learning (MIL) \cite{MIL} is a mainstream approach to solving weakly supervised object detection (WSOD). With the development of Deep Convolution Neural Network, some works \cite{WSDDN,MFMIL,ContextLocnet,PDA,WCCN} begin to utilize the framework combining MIL and CNN to resolve WSOD. Since the landmark multiple instance detection network (MIDN) model named WSDDN\cite{WSDDN} was proposed, weakly supervised object detection received more attention. 

Further, more refined instance selection methods based on OICR were put forward after OICR \cite{OICR} proposed multi-stage online instance classifier refinements sharply improving the detection performance of WSDDN. For instance, OAIL \cite{OAIL} considered a context classification loss to find  proposals covering the whole object and impose a spatial restriction on negative instances. OIM \cite{OIM} adopted a spatial graph and an appearance graph to mine less discriminative and missing instances. UI \cite{UI}  employed QR factorization to get an orthogonal matrix. The orthogonal matrix could make the parameters of each class in detection branches different and it could help the instance selection. Boosted OICR \cite{BoostOICR} proposed an adaptive supervision aggregation function to dynamically change the IoU for selecting instances, which alleviated the effect of selecting discriminative deformable instances.  SLV \cite{SLV} merged the average scores of filtered proposals to the pixels of an image and then minimized bounding rectangles of connected regions in the image by a thresholding method. AIR \cite{AIR} proposed an adaptive instance refinement framework to make the network mine accurate bounding boxes by the proposal scores. These methods focused on how to select better instances in the instance classifier refinements.  Furthermore, some works \cite{DNP,HRPRN,ZLDN,Pred_Net} based on other architectures instead of MIDN were also proposed to select better instances.

Most of the models mentioned above used a single MIDN. And some researchers think employing two or more MIDNs to complement each MIDN may be helpful to escape from the local minima. For example, D-MIL \cite{D-MIL} introduced discrepantly collaborative modules into the network, which utilized multiple MIL learners to complement each other’s learning. MGSC \cite{MGSC} proposed a pyramidal MIDN to tackle the local optima problem, which performed the proposal removal guided by the segmentation masks. And the segmentation masks were produced by some weakly supervised segmentation methods. C-MIDN \cite{C-MIDN} adopted a coupled complementary multiple instance detection network to remove proposals with the guidance of semantic segmentation derived from AffinityNet \cite{AffinityNet}. These methods paid attention to improving WSOD performance by complementing and removing instances among different MIDNs. Apart from these methods, some other means were also used to tackle WSOD, such as generating high-quality proposals \cite{W-RPN,PGPS}, backbone improvement \cite{EDRN}, curriculum learning \cite{MICL,CSPCL,MEFF}, reinforcement learning \cite{RSAL}, continuation optimization \cite{C-MIL}, generative adversarial learning \cite{GAL}, learnable bounding box adjusters \cite{LBBA}. In addition, extra pix-level information and segmentation tasks were introduced into some works \cite{WCCN,TS2C,SDCN,WS-JDS,CSC,WSOD2,OCR} to guide MIDN,  which improved the detection performance by using more complex network architectures.

Recently, several works \cite{DPS, OPG} focusing on the positive-negative imbalance problem were also proposed with similar purposes to ours. DPS \cite{DPS} utilized a dynamic proposal sampling strategy to collect purified positive instances and meanwhile mined hard negative instances.  OPG \cite{OPG} introduced an online active proposal set generation algorithm, which could dynamically use different proposal sampling methods to select proposals and part them to obtain an active proposal set for optimization. However, different from these methods, the OPIS method is more straightforward, which only uses progressive mixed sampling and reweights instances. And the training overheads of the proposed method are almost the same as OICR. Furthermore, the OPIS algorithm could also be easily integrated into other OICR-based methods.

\subsection{Sampling method for imbalance problem}

The foreground-background (i.e. positive-negative) imbalance problem is common in fully supervised object detection, which also exists in weakly supervised object detection. Hard sampling and soft sampling are commonly-used approaches to tackle the instance level imbalance problem. 

Random sampling was a classical sampling method in hard sampling. And recently some other hard sampling methods were proposed gradually. OHEM \cite{OHEM} utilized loss  of samples to select hard samples. However, it caused extra memory and low training speed. IoU-balanced sampling \cite{LIBRARCNN} employed the IoU interval to divide negative samples into $K$ bins and then randomly select the same number of samples in each bin to mine hard samples, which was an efficient and low-cost method.

Different from hard sampling, soft sampling usually adjusts the importance of each example. For instance, focal loss \cite{Focal_loss} assigned more weight to hard examples. Prime sample attention \cite{PISA} improved the importance of positive instances with higher IoUs. Meanwhile, negative instances possessing higher foreground classification scores were also enhanced.

During the training phase, weakly supervised object detection has no ground truth bounding box annotations. Therefore, the proposals can not be simply sorted by the loss. However, the IoU-balanced sampling is utilized without loss sorting, which inspires our work. In addition, the proposed method also reweights instances to improve positive instance importances, which belongs to soft sampling methods.

\section{Method}
\label{Method}

As mentioned above, this paper focuses on the balance of positive instances and negative instances to further improve the performance of WSOD. To this end, the proposed method is implemented on the classical OICR framework and the overall pipeline is shown in Fig. \ref{fig2}. First, an image with about 2000 unclassified instances (object proposals), generated by Selective Search, is fed into convolutional layers with the RoI-pooling layer for the extraction of fixed-size instance feature maps. Then the feature maps are fed into two fully connected layers to generate instance feature vectors. These instance feature vectors are branched into different streams: the multiple instance detection network and K  instance classifier refinements. The MIDN is supervised by the image-level label. The first instance classifier refinement is under the supervision of the MIDN. And the supervision of the $k^{th}$ instance classifier refinement depends on the $\{k-1\}^{th}$ instance classifier refinement. In this section, we will introduce the framework and the method in detail.

\subsection{Notations}
Before introducing the proposed method, some notations applied throught the paper will be summarized first. An image $\mathbf{X}$ with instances $\mathcal{R}$ and an image-level label $\mathbf{Y}=\left [y_1,y_2,\dots,y_C \right ]$ is fed into the network. $y_c=1$ or $0$ demonstrates the object of class $c$ is in the image or not. $C$ is the number of total object classes and $C+1$ means the background class. For $k^{th}$ refinement, $\phi_{cr}^{k}$ and $w_{r}^{k}$ are the predicted score and the loss weight of the instance $r$ belong to class c respectively. $P_c^k$ and $N_{c}^{k}$ are the sets of positive instances and negative instances after categorizing instances. Correspondly, $n_c^{Pk}$ and $n_c^{Nk}$ represent the number of instances in two sets. $P_{c}^{\prime k}$ and $N_{c}^{\prime k}$ contain positive instances and negative instances reselected respectively.   $I_{r}^{k}$ means the highest IoU gained by comparing the overlap between instance $r^k$ and the instance cluster centers. $T$ is a progressive parameter used in the phase of instance-balanced fine-tuing.  

\subsection{The overall training scheme}
\label{The overall training scheme}
The training process comprises two phases: normal training and fine-tuning for progressive instance balance. We denote that $T_0$ is the transition iteration from normal training to fine-tuning and $T_1$ is the final iteration. During the normal training phase, the network is trained following the top-scoring principle. And then instances belong to $P_c^k$ are reweighted by Eq. (\ref{eq12}).  During the fine-tuning for progressive instance balance, we apply an instance-balanced strategy and reweight positive instances by Eq. (\ref{eq13})  after selecting positive instances and negative instances. Moreover, we train the network under the final loss function as in Eq. (\ref{eq1}).
\begin{equation}
	\label{eq1}
	L=L_{MIDN}+\sum_{k=1}^{K}L_{OPIS}^{k},
\end{equation}
where the loss $L_{MIDN}$ is stated in later Eq. (\ref{eq2}) and the loss $L_{OPIS}^{k}$ is defined in later Eq. (\ref{eq5}) and Eq. (\ref{eq14}).

Obviously, the proposed method needs no extra network parameters. Furthermore, it could be readily used in other models based on OICR. The procedure is summarized in Algorithm \ref{alg1}.

\begin{algorithm}[h]
	\caption{The overall training procedure}
	\label{alg1}
	\begin{algorithmic}[1]
		\REQUIRE training dataset $D=(\mathbf{X}^{n},\mathbf{Y}^{n},\mathcal{R}^{n})$; refinement times $K$
		\ENSURE model $f(D,\theta)$
		\STATE {\textbf{Phase 1: Normal training}}
		\FOR{$t=1$ \textbf{to} $T_0$}
		\STATE {Sample mini-batch $D_t$ from $D$ and feed it into the network for generating instance scores for each branch. }
		\FOR{$k=1$ \textbf{to} $K$}
		\STATE {Obtain positive instances $P_c^{k}$ and negative instances $N_{c}^{k}$, see Section \ref{Progressive instance balance}. }
		\STATE {Reweighting positive instances belong to $P^{k}$ by Eq. (\ref{eq12}), see Section \ref{Progressive instance reweighting}.  }
		\ENDFOR
		\STATE {Compute the network loss $L$ by Eq. (\ref{eq1}), see Section \ref{The overall training scheme}.}
		\STATE {Update network parameters $\theta$. }
		\ENDFOR
		\STATE{\textbf{Phase 2: Fine-tuning for instance balance}}
		\FOR{$t=T_0+1$ \textbf{to} $T_1$}
		\STATE {Sample mini-batch $D_t$ from $D$ and feed it into the network for generating instance scores for each branch. }
		\FOR{$k=1$ \textbf{to} $K$}
		\STATE {Obtain positive instances $P_c^{k}$ and negative instances $N_{c}^{k}$, see Section \ref{Progressive instance balance}. }
		\FOR{$c=1$ \textbf{to} $C$}
		\STATE {Reselect negative instances from $N_{c}^{k}$ to $N_{c}^{\prime k}$ and positive instances form $P_{c}^{k}$ to $P_{c}^{\prime k}$, see Section \ref{Progressive instance balance}. }
		\ENDFOR
		\STATE {Reweighting positive instances belong to $P^{\prime k}$ by Eq. (\ref{eq13}), see Section \ref{Progressive instance reweighting}.  }
		\ENDFOR
		\STATE {Compute the network loss $L$ by Eq. (\ref{eq1}), see Section \ref{The overall training scheme}.}
		\STATE {Update network parameters $\theta$. }
		\ENDFOR
	\end{algorithmic}
	
\end{algorithm}

\subsection{Multiple instance detection network}
Many existing weakly supervised object detectors are based on a MIDN network. Here we employ the MIDN proposed by Bilen and Vedaldi, which used two data streams to acquire instance scores. The basic MIDN module is as follows. Given an image $\rm\mathbf{X}$ with the image-level label $\mathbf{Y}$ and instances $\mathcal{R}$, the final instance features extracted by CNN backbone and two fully connected layers can be obtained. Then two matrices $\rm\mathbf{x}^{cls}$,$\rm\mathbf{x}^{det}\in \mathbb{R} ^{C\times |\mathcal{R}|} $ are gained  after the instance features are fed into classification data stream and detection data stream, where $|\mathcal{R}|$ means the number of image instances. Subsequently, the matric $\rm\mathbf{x}^{cls}$ is passed through a softmax operation over classes to generate $[\sigma_{c}({\rm\mathbf{x}^{cls})}]_{ij}= \frac{e^{x^{cls}_{ij}}}{ \sum_{c=1}^{C}e^{x_{cj}^{cls}}} $  while the matric $\rm\mathbf{x}^{det}$ is passed through a softmax operation over instances to generate $[\sigma_{d}({\rm\mathbf{x}^{det})}]_{ij}= \frac{e^{x^{det}_{ij}}}{ \sum_{r=1}^{|\mathcal{R}|}e^{x_{ir}^{det}}}$. The score of each instance is obtained by element-wise product ${\rm\mathbf{x}}^{\mathcal{R}}=\sigma_{c}({\rm\mathbf{x}^{cls})} \odot  \sigma_{d}({\rm\mathbf{x}^{det})}$. Besides, the score on class $c$ at the image level is procured by $y^{\prime}_{c}=\sum_{r=1}^{|\mathcal{R}|}x_{cr}^{\mathcal{R}}$. Eventually, the loss function for training the basic MIDN module is as follows:
\begin{equation}
	\label{eq2}
	L_{MIDN}=-\sum_{c=1}^{C}\{y_c\log{y^{\prime}_{c}}+(1-y_c)\log{(1-y^{\prime}_{c})} \}.
\end{equation}

\subsection{Online progressive instance-balanced sampling}

\subsubsection{Progressive instance balance}
\label{Progressive instance balance}
As shown in Fig. \ref{fig2}, for the $k^{th}$ refinement, the label $\mathbf{Y}$ and instance scores $\mathbf{\phi}^{k-1}$ of the $\{k-1\}^{th}$ refinement are used to obtain the supervision $\mathcal{S}^{k}$. During normal training, the instance cluster center of each class is obtained according to the top-scoring principle for the first step. Following \cite{OICR}, highly spatially adjacent proposals should be assigned the same label. So other positive instances and negative instances are selected by the principle of spatial similarity, i.e. IoU. Each instance is assigned the label and weight decided by the instance cluster center. The instance cluster center $\mathcal{I}_{c}^{k}$ of the class $c$ is gained by Eq. (\ref{eq3}).
\begin{equation}
	\label{eq3}
	\mathcal{I}_{c}^{k}=\arg \max_{r}\phi^{k-1}_{cr}.
\end{equation}

Assume an instance cluster center $\mathcal{I}_{c}^{k}$ has an adjacent instance $r$, if the IoU $I_{r}^{k}$ is greater than or equal to $\lambda_{ng}$, we set the instance label $y_{cr}^{k}=1$ and $y_{c^{\prime}r}^{k}=0$, where $c^{\prime}\neq c$. And if the IoU is less than $\lambda_{ig}$, the instance is ignored and its weight is set to zero. Otherwise, the instance $r$ is labeled as a negative instance, which means $y_{(C+1)r}^{k}=1$ and $y_{c^{\prime\prime} r}=0$, $c^{\prime\prime} \neq C+1$. For positive instances and negative instances, their weights are equal to the score of instance cluster center $\mathcal{I}_{c}^{k}$ in the $\{k-1\}^{th}$ branch  as in Eq. (\ref{eq4}). 
\begin{equation}
	\label{eq4}
	w_{r}^{k}=\phi^{k-1}_{c\mathcal{I}_c^{k}},
\end{equation}
where $w_{r}^{k}$ means the weight of instance $r^{k}$ for the $k^{th}$ refinement and $\phi^{k-1}_{c\mathcal{I}_c^{k}}$ denotes the score of instance cluster center $\mathcal{I}_{c}^{k}$ in the $\{k-1\}^{th}$ branch.

The loss function of $k^{th}$ refinement is in Eq.(\ref{eq5}):
\begin{equation}
	\label{eq5}
	L_{OPIS}^{k}=-\frac{1}{|\mathcal{R}|}\sum_{r=1}^{|\mathcal{R}|}\sum_{c=1}^{C+1}w_{r}^{k}y^{k}_{cr}\log \phi_{cr}^{k}.
\end{equation}

During the fine-tuning phase, we further reselect negative instances. $T$ is defined as a progressive parameter:
\begin{equation}
	\label{eq6}
	T=\frac{T_{n}-T_{0}}{T_1-T_{0}},
\end{equation}
where $T_n$ denotes the current training iteration, $T_0$ means the iteration that starts to apply the progressive instance balance, and $T_1$ is the total training iteration. In addition, a ratio of negative instances to positive instances for each class is 
\begin{equation}
	\label{eq7}
	\mu=\mu_{s}-(\mu_{s}-4)T,
\end{equation}
where $\mu_{s}$ represents an initial sampling ratio.

Instances gathered by the instance cluster center $\mathcal{I}_{c}^{k}$ could be divied into $P_{c}^{k}$ and $N_{c}^{k}$. And $n_{c}^{\prime Nk}=\lfloor \mu n_{c}^{Pk} \rfloor$ negative instances are needed to be reselected. Inspired by IoU-balanced sampling, we split $(\lambda_{ig},\lambda_{ng})$ equally into four bins ( e.g., $\lambda_{ig}=0.1$ and $\lambda_{ng}=0.5$). Different from IoU-balanced sampling, only $n_{c}^{Pk}$ negative instances are selected by the uniform sampling in the bin $j$, which has $B_{cj}$ negative instances. The instances in bin $j$ are selected by the probability as in Eq. (\ref{eq8}).
\begin{equation}
	\label{eq8}
	p_{cj}=\frac{n_{c}^{Pk}}{B_{cj}},\ j\in [1,4].
\end{equation}

After that, we randomly sample the remaining required negative instances by the probability as in Eq. (\ref{eq9}).
\begin{equation}
	\label{eq9}
	p_{c}=\frac{n_{ c}^{\prime Nk}-4n_{c}^{Pk}}{n_{c}^{Nk}-4n_{c}^{Pk}}.
\end{equation}

And these instances are added into $N_{c}^{\prime k}$. Unlike instance-level supervisions, image-level supervisions do not provide ground-truth boxes. The location of the instance cluster center gained by the top-scoring principle is not accurate. The instance may only locate parts of an object. Directly using IoU-balanced sampling maybe guide the distribution of negative instances close to  one of the negative instances located in parts of objects. Whereas, the instance cluster center could gradually cover most areas of objects with the training going on. And the proportion of instances selected by IoU-balanced sampling increase gradually, which could guide the distribution of negative instances close to a better one of hard negative instances. And the effectiveness of progressive IoU-balanced sampling is demonstrated in detailed ablation studies.  

In addition,  we find that sometimes the instance cluster center $\mathcal{I}_c^{k}$ has no negative instance.  Meanwhile, it has many positive instances $\mathcal{P}_c^{k}$ with low scores.  Obviously, these positive instances may locate in very small objects or small parts of objects. Slight offsets between the instance cluster center and positive instances make positive instance scores shrink dramatically. Although the corresponding weights of these instances are also low, they will introduce non-negligible noises, which may make the network fall into local minima,  after progressive instance reweighting during the whole training. Therefore, we apply neglect of these positive instances and only keep the instance cluster. We set a threshold $I_t$ by Eq. (\ref{eq10}) to reselect positive instances. 
\begin{equation}
	\label{eq10}
	I_t=I_0+\alpha\frac{T_n}{T_1},
\end{equation}
where $\alpha$ is a set value and $I_0$ is usually set as 0.05. Then if the sum of the scores of positive instances is less than $I_t$, only the instance cluster center $\mathcal{I}_c^{k}$  is added into $P^{\prime k}_{c}$. Otherwise, all positive instances are added into $P^{\prime k}_{c}$, i.e. $P^{\prime k}_{c}=P^{k}_{c}$.

Moreover, the weights of instances in Eq. (\ref{eq4})  are changed, as in Eq. (\ref{eq11}).
\begin{equation}
	\label{eq11}
	{w_{r}^{k}} = \begin{cases}
		\phi^{k-1}_{c\mathcal{I}_c^{k}}, &{\text{for} }\ {r \in} (P_{c}^{\prime k},N_{c}^{\prime k}) \\ 
		{0}, &{\text{otherwise}}.
	\end{cases}
\end{equation}

\subsubsection{Progressive instance reweighting}
\label{Progressive instance reweighting}
As mentioned above, the number of negative instances is far more than positive instances during the training stage in OICR. However, it is unreasonable to uniformly set the selected instance weight as the score of the highest score proposal for each class in OICR, which causes those positive proposals to be drowned out by negative proposals. Simply, we could think of balancing positive instances and negative instances by reducing the number of selected negative instances. But during the process of training, due to no instance-level annotations provided, classifiers have no strong discriminative capability and the model cannot accurately find suitable negative instances to compute the loss. Meanwhile, reducing the number of selected negative instances prematurely may introduce noise into backpropagation by picking low-quality negative instances. Therefore, during the training stage, the model should equally focus on positive instances and negative instances by reweighting positive instance weights.

In addition, in the absence of ground-truth boxes, the surrounding background of the instance cluster center is not sure because the instance cluster center may only locate at part of the object. Therefore, equally assigning instance weights or only using IoU to reweight will introduce noises into the network by low-scoring positive instances. Furthermore, the top-scoring instances of different classes are only judged by the classifier score matrice of the $\{k-1\}^{th}$ branch in the $k^{th}$  supervision of the instance classifier refinement. However, the $k^{th}$ classifier also has classification ability with the training going on. Based on the idea "many could better than one", the score matrice of the $k^{th}$ classifier is introduced into the $k^{th}$ supervision for reweighting the weights of positive instances $P_c^{\prime k}$ as in Eq. (\ref{eq12}).

\begin{equation}
	\label{eq12}
	w_{r}^{k}=(\beta e^{\phi_{cr}^{k}}+(1-\beta)e^{I_{r}})\phi^{k-1}_{c\mathcal{I}_c^{k}},
\end{equation}
where $w_{r}^{k}$ is the initial weight of positive instance $r$ in the $k^{th}$ branch,  $\phi_{cr}^{k}$ is the score of instance $r$ in the $k^{th}$ branch for class $c$, $I_{r}$ is the IoU between instance $r$ and the highest-scoring proposal $\mathcal{I}_{c}^{k-1}$, and $\beta$ is the hyper parameter used for balancing $\phi_{cr}^{k}$ and $I_{r}$.

During the fine-tuning phase, the number of selected negative proposals begins to decrease. And the weights of positive proposals need to be reduced. Otherwise, the model will focus too much on positive instances, leading to the part domination problem. So we change the reweighting function in Eq. (\ref{eq12}) to an attenuated version, as in Eq. (\ref{eq13}).

\begin{equation}
	\label{eq13}
	w_{r}^{k}=e^{-\gamma T}(\beta e^{\phi_{cr}^{k}}+(1-\beta)e^{I_{r}^{}})\phi^{k-1}_{c\mathcal{I}_c^{k}},
\end{equation}
where $\gamma$ is used to control the drop of positive proposal weights.

Compared to the origin instances in an image, the number of instances decreases a lot after reselecting instances. Therefore, the loss function of $k^{th}$ refinement should be changed as in Eq. (\ref{eq14}).
\begin{equation}
	\label{eq14}
	L_{OPIS}^{k}=-\frac{1}{|\mathcal{R}|}\sum_{r=1}^{|\mathcal{R}|}\sum_{c=1}^{C+1}\zeta w_{r}^{k}y^{k}_{cr}\log \phi_{cr}^{k},
\end{equation}

\begin{equation}
	\label{eq15}
	{\zeta} = \begin{cases}
		{1}, & t \leq T_0 \\ 
		{\frac{|\mathcal{R}|}{|\mathcal{R}_s|}}, &{T_0} < t \leq T_1 
	\end{cases},
\end{equation}
where $\mathcal{R}_s=(P_{c}^{\prime k},N_{c}^{\prime k})$ and  $|\mathcal{R}_s|$ is the number of instances in $\mathcal{R}_s$. Finally, the above steps are summarized in Algorithm \ref{alg2}.

\begin{algorithm}[ht]
	
	\caption{Online progressive instance-balanced sampling ( for fine-tuning phase)}
	\label{alg2}
	\begin{algorithmic}[1]
		\REQUIRE Image label $\mathbf{Y}$; instance score $\phi^{k-1}$ and $\phi^{k}$; instances $\mathcal{R}$.
		\ENSURE Instance label $Y^k = \{y^k_r\}_{r=1}^{\mathcal{R}}$, $y^k_r=[y_{1r}^{k},\dots,y_{(C+1)r}^{k}]^{T}$; instance weights  $w_r^{k}$, where $r$ $\in$ $\mathcal{R}$.
		\STATE Initialize all elements in $y_r^k$ to 0.
		\FOR{$c=1$ $\textbf{to}$ $C$}
		\IF {$y_c=1$}
		\STATE {Get an instance cluster center $\mathcal{I}_{c}^{k}$ by Eq. (\ref{eq3}).}
		\ENDIF
		\ENDFOR
		\FOR {$r=1$ $\textbf{to}$ $\mathcal{R}$}
		\STATE Compute the highest IoU $I_r^k$ and gain the class $c$.
		\IF {$I_r^k$ $\geq$ $\lambda_{ng}$}
		\STATE {$y_{cr}^k=1$ and $w_r^k=\phi^{k-1}_{c\mathcal{I}_c^{k}}$.}
		\ELSIF {$I_r^k$ $\leq$ $\lambda_{ig}$}
		\STATE {$w_r^k=0$.}
		\ELSE
		\STATE {$y_{(C+1)r}=1$ and $w_r^k=\phi^{k-1}_{c\mathcal{I}_c^{k}}$.}
		\ENDIF
		\ENDFOR
		\FOR{$c=1$ $\textbf{to}$ $C$}
		\IF {$n_{c}^{Nk}$ $>$ $0$}
		\STATE {Compute the number of negative instances needed to reselect $n_{c}^{\prime Nk}=\lfloor \mu n_{c}^{Pk}$ $\rfloor$.}
		\STATE {Gain reselected negative instances $N_{c}^{\prime k}$  by Eq. (\ref{eq8}) and (\ref{eq9}).}
		\ELSE
		\STATE {Gain reselected positive instances $P_{c}^{\prime k}$ by the threshold in Eq. (\ref{eq10}).}
		\ENDIF
		\ENDFOR
		\FOR {r $\in$ $P^{\prime k}$}
		\STATE {Compute $w_{r}^{k}$ by Eq. (\ref{eq13}).}
		\ENDFOR
	\end{algorithmic}
	
\end{algorithm}

\begin{table*}[ht]
	\centering
	\caption{Detection average precision ($\%$) for different methods on the PASCAL VOC 2007 test set.}
	\label{table1}
	\resizebox{\textwidth}{39mm}{
		\begin{tabular}{lccccccccccccccccccccc}
			\toprule
			Method & aero  & bike  & bird  & boat  & bottle & bus   & car   & cat   & chair & cow   & table & dog   & horse & mbike & person & plant & sheep & sofa  & train & tv    & mAP \\
			\midrule
			WSDDN \cite{WSDDN} & 46.4  & 58.3  & 35.5  & 25.9  & 14.0  & 66.7  & 53.0  & 39.2  & 8.9   & 41.8  & 26.6  & 38.6  & 44.7  & 59.0  & 10.8  & 17.3  & 40.7  & 49.6  & 56.9  & 50.8  & 39.3  \\
			OICR \cite{OICR}  & 58.0  & 62.4  & 31.1  & 19.4  & 13.0  & 65.1  & 62.2  & 28.4  & 24.8  & 44.7  & 30.6  & 25.3  & 37.8  & 65.5  & 15.7  & 24.1  & 41.7  & 46.9  & 64.3  & 62.6  & 41.2  \\
			PCL \cite{PCL}   & 54.4  & 69.0  & 39.3  & 19.2  & 15.7  & 62.9  & 64.4  & 30.0  & 25.1  & 52.5  & 44.4  & 19.6  & 39.3  & 67.7  & 17.8  & 22.9  & 46.6  & 57.5  & 58.6  & 63.0  & 43.5  \\
			TS2C \cite{TS2C}  & 59.3  & 57.5  & 43.7  & 27.3  & 13.5  & 63.9  & 61.7  & 59.9  & 24.1  & 46.9  & 36.7  & 45.6  & 39.9  & 62.6  & 10.3  & 23.6  & 41.7  & 52.4  & 58.7  & 56.6  & 44.3  \\
			WSRPN \cite{RegionProposal} & 57.9  & 70.5  & 37.8  & 5.7   & 21.0  & 66.1  & 69.2  & 59.4  & 3.4   & 57.1  & \textbf{57.3}  & 35.2  & 64.2  & 68.6  & \textbf{32.8}  & \textbf{28.6}  & 50.8  & 49.5  & 41.1  & 30.0  & 45.3  \\
			WS-JDS \cite{WS-JDS} & 52.0  & 64.5  & 45.5  & 26.7  & 27.9  & 60.5  & 47.8  & 59.7  & 13.0  & 50.4  & 46.4  & 56.3  & 49.6  & 60.7  & 25.4  & 28.2  & 50.0  & 51.4  & 66.5  & 29.7  & 45.6  \\
			C-SPCL \cite{CSPCL} & 63.4  & 55.0  & 52.8  & \textbf{36.6}  & 10.7  & 66.3  & 57.0  & 69.5  & 7.2   & 52.5  & 14.4  & \textbf{64.6}  & \textbf{69.4}  & 57.7  & 28.4  & 15.8  & 43.7  & 42.3  & 69.3  & 40.5  & 45.9  \\
			Adapt-WSL \cite{Apative_WSL} & 56.9  & 65.4  & 43.9  & 24.8  & 19.5  & \textbf{71.1}  & 59.6  & 54.7  & 23.4  & 49.1  & 45.6  & 57.8  & 45.8  & 63.0  & 12.9  & 22.5  & 36.0  & 51.1  & 64.4  & 56.9  & 46.2  \\
			C-WSL \cite{C_WSL} & 62.9  & 64.8  & 39.8  & 28.1  & 16.4  & 69.5  & 68.2  & 47.0  & \textbf{27.9}  & 55.8  & 43.7  & 31.2  & 43.8  & 65.0  & 10.9  & 26.1  & 52.7  & 55.3  & 60.2  & 66.6  & 46.8  \\
			MELM \cite{MELM} & 55.6  & 66.9  & 34.2  & 29.1  & 16.4  & 68.8  & 68.1  & 43.0  & 25.0  & 65.6  & 45.3  & 53.2  & 49.6  & 68.6  & 2.0   & 25.4  & 52.5  & 56.8  & 62.1  & 57.1  & 47.3  \\
			OAIL \cite{OAIL}  & 61.5  & 64.8  & 43.7  & 26.4  & 17.1  & 67.4  & 62.4  & 67.8  & 25.4  & 51.0  & 33.7  & 47.6  & 51.2  & 65.2  & 19.3  & 24.4  & 44.6  & 54.1  & 65.6  & 59.5  & 47.6  \\
			Boosted OICR \cite{BoostOICR} & \textbf{68.6}  & 62.4  & \textbf{55.5}  & 27.2  & 21.4  & \textbf{71.1}  & 71.6  & 56.7  & 24.7  & 60.3  & 47.4  & 56.1  & 46.4  & 69.2  & 2.7   & 22.9  & 41.5  & 47.7  & \textbf{71.1}  & \textbf{69.8}  & 49.7  \\
			OPG \cite{OPG}  & 63.0  & 65.3  & 49.2  & 31.7  & 25.3  & 70.9  & 70.9  & 58.1  & 27.4  & 58.6  & 44.7  & 47.0  & 47.2  & 69.8  & 13.1  & 26.1  & 49.9  & 51.8  & 61.7  & 68.2  & 50.0  \\
			OIM \cite{OIM}  & 55.6  & 67.0  & 45.8  & 27.9  & 21.1  & 69.0  & 68.3  & \textbf{70.5}  & 21.3  & 60.2  & 40.3  & 54.5  & 56.5  & 70.1  & 12.5  & 25.0  & 52.9  & 55.2  & 65.0  & 63.7  & 50.1  \\
			C-MIL \cite{C-MIL} & 62.5  & 58.4  & 49.5  & 32.1  & 19.8  & 70.5  & 66.1  & 63.4  & 20.0  & 60.5  & 52.9  & 53.5  & 57.4  & 68.9  & 8.4   & 24.6  & 51.8  & 58.7  & 66.7  & 63.5  & 50.5  \\
			DPS \cite{DPS}   & 60.1  & \textbf{74.5}  & 51.9  & 29.6  & 30.2  & 68.8  & \textbf{72.6}  & 44.6  & 19.8  & \textbf{66.0}  & 48.8  & 43.7  & 63.2  & 68.2  & 17.7  & 25.1  & 53.7  & \textbf{60.8}  & 56.1  & 63.1  & 50.9  \\
			Baseline & 63.2  & 69.5  & 51.4  & 14.2  & 18.7  & 68.4  & 70.4  & 40.4  & 22.9  & 57.3  & 46.1  & 45.8  & 50.7  & \textbf{70.7}  & 13.7  & 26.2  & 47.9  & 55.6  & 57.8  & 67.7  & 47.9  \\
			OPIS (ours) & 65.4  & 72.6  & 54.8  & 22.1  & \textbf{30.5}  & 70.1  & 71.7  & 56.3  & 26.0  & 63.9  & 46.2  & 53.1  & 47.6  & 70.2  & 13.8  & 20.8  & \textbf{54.3}  & 51.3  & 68.6  & 68.6  & \textbf{51.4}  \\
			\midrule
			OICR-Ens.+FRCNN \cite{OICR} & 65.6  & 67.2  & 47.2  & 21.6  & 22.1  & 68.0  & 68.5  & 35.9  & 5.7   & 63.1  & 49.5  & 30.3  & 64.7  & 66.1  & 13.0  & 25.6  & 50.0  & 57.1  & 60.2  & 59.0  & 47.0  \\
			TS2C+FRCNN \cite{TS2C} & -     & -     & -     & -     & -     & -     & -     & -     & -     & -     & -     & -     & -     & -     & -     & -     & -     & -     & -     & -     & 48.0  \\
			C-WSL+FRCNN \cite{C_WSL} & 62.9  & 68.3  & 52.9  & 25.8  & 16.5  & 71.1  & 69.5  & 48.2  & 26.0  & 58.6  & 44.5  & 28.2  & 49.6  & 66.4  & 10.2  & 26.4  & 55.3  & 59.9  & 61.6  & 62.2  & 48.2  \\
			PCL-Ens.+FRCNN \cite{PCL} & 63.2  & 69.9  & 47.9  & 22.6  & 27.3  & 71.0  & 69.1  & 49.6  & 12.0  & 60.1  & 51.5  & 37.3  & 63.3  & 63.9  & 15.8  & 23.6  & 48.8  & 55.3  & 61.2  & 62.1  & 48.8  \\
			Boosted OICR+FRCNN \cite{BoostOICR} & \textbf{65.8}  & 58.6  & 55.0  & \textbf{32.4}  & 19.5  & \textbf{74.2}  & 71.4  & 70.9  & 19.2  & 54.8  & 46.2  & 67.5  & 57.0  & 65.6  & 1.4   & 16.7  & 40.4  & 53.0  & 69.5  & 61.1  & 50.0  \\
			WSRPN-Ens.+FRCNN \cite{RegionProposal} & 63.0  & 69.7  & 40.8  & 11.6  & 27.7  & 70.5  & 74.1  & 58.5  & 10.0  & 66.7  & \textbf{60.6}  & 34.7  & 75.7  & 70.3  & 25.7  & 26.5  & 55.4  & 56.4  & 55.5  & 54.9  & 50.4  \\
			MIL-PCL+GAM+REG \cite{E2E} & 57.6  & 70.8  & 50.7  & 28.3  & 27.2  & 72.5  & 69.1  & 65.0  & \textbf{26.9}  & 64.5  & 47.4  & 47.7  & 53.5  & 66.9  & 13.7  & 29.3  & 56.0  & 54.9  & 63.4  & 65.2  & 51.5  \\
			WS-JDS+FRCNN \cite{WS-JDS} & 64.8  & 70.7  & 51.5  & 25.1  & \textbf{29.0}  & 74.1  & 69.7  & 69.6  & 12.7  & \textbf{69.5}  & 43.9  & 54.9  & 39.3  & \textbf{71.3}  & \textbf{32.6}  & \textbf{29.8} & 57.0  & 61.0  & 66.6  & 57.4  & 52.5  \\
			OIM+FRCNN \cite{OIM} & 53.4  & 72.0  & 51.4  & 26.0  & 27.7  & 69.8  & 69.7  & \textbf{74.8}  & 21.4  & 67.1  & 45.7  & 63.7  & 63.7  & 67.4  & 10.9  & 25.3  & 53.5  & 60.4  & \textbf{70.8}  & 58.1  & 52.6  \\
			C-MIL+FRCNN \cite{C-MIL} & 61.8  & 60.9  & 56.2  & 28.9  & 18.9  & 68.2  & 69.6  & 71.4  & 18.5  & 64.3  & 57.2  & \textbf{66.9}  & \textbf{65.9}  & 65.7  & 13.8  & 22.9  & 54.1  & \textbf{61.9}  & 68.2  & 66.1  & 53.1  \\
			OPIS+REG (ours) & 59.6  & \textbf{77.3}  & \textbf{59.0}  & 13.0  & 22.4  & 71.4  & \textbf{73.9}  & 62.5  & 26.4  & 66.7  & 47.9  & 53.2  & 63.7  & 69.8  & 20.9  & 26.6  & \textbf{63.1}  & 57.4  & 67.6  & \textbf{70.8}  & \textbf{53.2}  \\
			\bottomrule 
		\end{tabular}
	}
	
\end{table*}

\begin{table*}[ht]
	\centering
	\caption{Detection CorLoc ($\%$) for different methods on the PASCAL VOC 2007 trainval set.}
	\label{table2}
	\resizebox{\textwidth}{36mm}{
		\begin{tabular}{lccccccccccccccccccccc}
			\toprule
			Method & aero  & bike  & bird  & boat  & bottle & bus   & car   & cat   & chair & cow   & table & dog   & horse & mbike & person & plant & sheep & sofa  & train & tv    & CorLoc \\
			\midrule
			WSDDN \cite{WSDDN} & 68.9  & 68.7  & 65.2  & 42.5  & 40.6  & 72.6  & 75.2  & 53.7  & 29.7  & 68.1  & 33.5  & 45.6  & 65.9  & 86.1  & 27.5  & 44.9  & 76.0  & 62.4  & 66.3  & 66.8  & 58.0  \\
			OICR \cite{OICR} & 81.7  & 80.4  & 48.7  & 49.5  & 32.8  & 81.7  & 85.4  & 40.1  & 40.6  & 79.5  & 35.7  & 33.7  & 60.5  & 88.8  & 21.8  & 57.9  & 76.3  & 59.9  & 75.3  & 81.4  & 60.6  \\
			PCL  \cite{PCL}  & 79.6  & 85.5  & 62.2  & 47.9  & 37.0  & \textbf{83.8}  & 83.4  & 43.0  & 38.3  & 80.1  & 50.6  & 30.9  & 57.8  & 90.8  & 27.0  & 58.2  & 75.3  & \textbf{68.5}  & 75.7  & 78.9  & 62.7  \\
			TS2C \cite{TS2C} & 84.2  & 74.1  & 61.3  & 52.1  & 32.1  & 76.7  & 82.9  & 66.6  & 42.3  & 70.6  & 39.5  & 57.0  & 61.2  & 88.4  & 9.3   & 54.6  & 72.2  & 60.0  & 65.0  & 70.3  & 61.0  \\
			WSRPN \cite{RegionProposal} & 77.5  & 81.2  & 55.3  & 19.7  & 44.3  & 80.2  & 86.6  & 69.5  & 10.1  & \textbf{87.7}  & \textbf{68.4}  & 52.1  & \textbf{84.4}  & \textbf{91.6}  & \textbf{57.4}  & 63.4  & 77.3  & 58.1  & 57.0  & 53.8  & 63.8  \\
			WS-JDS \cite{WS-JDS} & 82.9  & 74.0  & \textbf{73.4}  & 47.1  & 60.9  & 80.4  & 77.5  & \textbf{78.8}  & 18.6  & 70.0  & 56.7  & 67.0  & 64.5  & 84.0  & 47.0  & 50.1  & 71.9  & 57.6  & \textbf{83.3}  & 43.5  & 64.5  \\
			C-SPCL \cite{CSPCL} & 63.4  & 55.0  & 52.8  & 36.6  & 10.7  & 66.3  & 57.0  & 69.5  & 7.2   & 52.5  & 14.4  & 64.6  & 69.4  & 57.7  & 28.4  & 15.8  & 43.7  & 42.3  & 69.3  & 40.5  & 45.9  \\
			Adapt-WSL \cite{Apative_WSL} & 80.8  & 78.8  & 72.1  & 50.5  & 43.1  & 77.2  & 83.8  & 65.4  & 38.3  & 82.2  & 40.7  & 70.0  & 75.5  & 88.0  & 16.7  & 48.0  & 72.2  & 56.2  & 73.4  & 69.2  & 64.1  \\
			C-WSL \cite{C_WSL} & 85.8  & 81.2  & 64.9  & 50.5  & 32.1  & 84.3  & 85.9  & 54.7  & 43.4  & 80.1  & 42.2  & 42.6  & 60.5  & 90.4  & 13.7  & 57.5  & \textbf{82.5}  & 61.8  & 74.1  & 82.4  & 63.5  \\
			MELM \cite{MELM} & -     & -     & -     & -     & -     & -     & -     & -     & -     & -     & -     & -     & -     & -     & -     & -     & -     & -     & -     & -     & 61.4  \\
			OAIL \cite{OAIL}   & 85.5  & 79.6  & 68.1  & 55.1  & 33.6  & 83.5  & 83.1  & 78.5  & 42.7  & 79.8  & 37.8  & 61.5  & 74.4  & 88.6  & 32.6  & 55.7  & 77.9  & 63.7  & 78.4  & 74.1  & 66.7  \\
			Boosted OICR \cite{BoostOICR} & \textbf{86.7}  & 73.3  & 72.4  & \textbf{55.3}  & 46.9  & 83.2  & 87.5  & 64.5  & \textbf{44.6}  & 76.7  & 46.4  & \textbf{70.9}  & 67.0  & 88.0  & 9.6   & 56.4  & 69.1  & 52.4  & 79.8  & 82.8  & 65.7  \\
			OPG \cite{OPG}  & 83.3  & 77.7  & 64.0  & 50.0  & 43.1  & 80.7  & 87.5  & 61.6  & 40.7  & 79.5  & 43.4  & 60.0  & 71.4  & 90.8  & 15.9  & \textbf{58.6}  & 76.3  & 62.6  & 74.9  & 83.2  & 65.3  \\
			OIM \cite{OIM}  & -     & -     & -     & -     & -     & -     & -     & -     & -     & -     & -     & -     & -     & -     & -     & -     & -     & -     & -     & -     & \textbf{67.2}  \\
			C-MIL \cite{C-MIL} & -     & -     & -     & -     & -     & -     & -     & -     & -     & -     & -     & -     & -     & -     & -     & -     & -     & -     & -     & -     & 65.0  \\
			DPS \cite{DPS}  & 81.8  & 86.1  & 67.1  & 51.5  & \textbf{53.3}  & 76.1  & 87.3  & 53.0  & 37.7  & 81.1  & 47.4  & 55.7  & 73.1  & 90.2  & 25.4  & 57.8  & 82.4  & 67.7  & 71.3  & \textbf{84.5}  & 66.5  \\
			Baseline & 83.3  & 83.1  & 67.3  & 42.0  & 46.2  & 80.2  & 88.0  & 53.8  & 39.9  & 80.8  & 46.0  & 54.2  & 70.7  & 89.6  & 16.6  & 49.5  & 70.1  & 58.1  & 71.5  & 83.2  & 63.7  \\
			OPIS (ours) & 84.6  & \textbf{85.9}  & 73.3  & 50.0  & 46.2  & 80.2  & \textbf{88.7}  & 56.1  & 44.2  & 81.5  & 47.9  & 65.3  & 71.8  & 90.4  & 14.8  & 54.2  & 77.3  & 57.5  & 74.1  & 82.8  & 66.3  \\
			\midrule
			OICR-Ens.+FRCNN \cite{OICR} & 85.8  & 82.7  & 62.8  & 45.2  & 43.5  & \textbf{84.8}  & 87.0  & 46.8  & 15.7  & 82.2  & 51.0  & 45.6  & 83.7  & 91.2  & 22.2  & 59.7  & 75.3  & 65.1  & 76.8  & 78.1  & 64.3  \\
			TS2C+FRCNN \cite{TS2C} & 84.2  & 74.1  & 61.3  & 52.1  & 32.1  & 76.7  & 82.9  & 66.6  & 42.3  & 70.6  & 39.5  & 57.0  & 61.2  & 88.4  & 9.3   & 54.6  & 72.2  & 60.0  & 65.0  & 70.3  & 61.0  \\
			C-WSL+FRCNN \cite{C_WSL} & \textbf{87.5}  & 81.6  & 65.5  & 52.1  & 37.4  & 83.8  & 87.9  & 57.6  & \textbf{50.3}  & 80.8  & 44.9  & 44.4  & 65.6  & 92.8  & 14.9  & \textbf{61.2}  & \textbf{83.5}  & 68.5  & 77.6  & \textbf{83.5}  & 66.1  \\
			PCL-Ens.+FRCNN \cite{PCL} & 83.8  & 85.1  & 65.5  & 43.1  & 50.8  & 83.2  & 85.3  & 59.3  & 28.5  & 82.2  & 57.4  & 50.7  & 85.0  & 92.0  & 27.9  & 54.2  & 72.2  & 65.9  & 77.6  & 82.1  & 66.6  \\
			WSRPN-Ens.+FRCNN \cite{RegionProposal} & 83.8  & 82.7  & 60.7  & 35.1  & 53.8  & 82.7  & 88.6  & 67.4  & 22.0  & \textbf{86.3}  & \textbf{68.8}  & 50.9  & \textbf{90.8}  & \textbf{93.6}  & 44.0  & \textbf{61.2}  & 82.5  & 65.9  & 71.1  & 76.7  & 68.4  \\
			MIL-PCL+GAM+REG \cite{E2E} & 80.0  & 83.9  & \textbf{74.2}  & \textbf{53.2}  & 48.5  & 82.7  & 86.2  & 69.5  & 39.3  & 82.9  & 53.6  & 61.4  & 72.4  & 91.2  & 22.4  & 57.5  & \textbf{83.5}  & 64.8  & 75.7  & 77.1  & 68.0  \\
			WS-JDS+FRCNN \cite{WS-JDS} & 79.8  & 84.0  & 68.3  & 40.2  & \textbf{61.5}  & 80.5  & 85.8  & \textbf{75.8}  & 29.7  & 77.7  & 49.5  & \textbf{67.4}  & 58.6  & 87.4  & \textbf{66.2}  & 46.6  & 78.5  & \textbf{73.7}  & \textbf{84.5}  & 72.8  & 68.6  \\
			OIM+FRCNN \cite{OIM} & -     & -     & -     & -     & -     & -     & -     & -     & -     & -     & -     & -     & -     & -     & -     & -     & -     & -     & -     & -     & \textbf{68.8}  \\
			OPIS+REG (ours) & 84.2  & \textbf{85.9}  & 68.8  & 43.1  & 35.9  & \textbf{84.8}  & \textbf{89.1}  & 66.0  & 41.3  & 84.2  & 49.4  & 62.8  & 82.3  & 93.2  & 35.0  & 60.8  & 78.4  & 59.4  & 79.8  & 82.8  & 68.4  \\
			\bottomrule
		\end{tabular}
	}
	
\end{table*}

\section{Experiment}
\subsection{Datasets and evaluation metrics}
The proposed method was evaluated on two datasets, which are PASCAL VOC 2007 and 2012 \cite{PASCAL_VOC}. These two datasets both contain 20 object categories, which are widely used for WSOD performance evaluation. We train the model on the $trainval$ sets following the standard practice in previous work. And only classification annotations are employed to train the model. For evaluation, mean Average Precision (mAP) and Correct Localization (CorLoc) are utilized to validate the effectiveness of the proposed method. First, mAP is used on the $test$ sets. Second, CorLoc is performed on the $trainval$ sets to evaluate the localization accuracy. Both metrics abide by the PASCAL VOC criterion that a predicted bounding box is deemed positive if the $IoU>$0.5 between a ground-truth bounding box and it.

\begin{table}[ht]
	\caption{Ablation study of OPIS main configurations. AT denotes the attenuated version of progressive instance reweighting.}
	\label{table3}
	\begin{center}
		\begin{tabular}{lc}
			\toprule
			Method & mAP \\
			\midrule
			Baseline &	47.9 \\
			+PIB & 50.4 \\
			+PIR w/o AT &  50.6    \\
			+PIB+PIR w/o AT & 48.4 \\
			+PIB+PIR  & $\textbf{51.4}$ \\
			\bottomrule 
		\end{tabular}
		
	\end{center}
\end{table}

\subsection{Implementation details}

For a fair comparison with other methods, VGG-16 model pre-trained on ImageNet dataset is chosen as the backbone of the basic MID module. However, we use the RoI-pooling layer instead of the SPP layer in the VGG-16. Following \cite{OICR}, we set K=3 times instance classifier refinements. The number of proposals per image is about 2000 generated by Selective Search.

The momentum and weight decay are set to 0.9 and 0.0005 respectively. During training, the mini-batch size for SGD is set to 2. The learning rate is initialized as 0.01 for the first 70K and 150K iterations for the PASCAL VOC 2007 and 2012 respectively. and then it decreases to 0.001 in the following 20K and 30K iterations for the PASCAL VOC 2007 and 2012 correspondingly, i.e. model is trained for about 18 epochs on VOC 2007 and 2012. For data augmentation, we take five image scales $\{$480, 576, 688, 864, 1200$\}$ and a horizontal flipping strategy during training and testing. We randomly choose one of five scales to resize the image and the longer side is resized into the selected scale. During testing, every image is augmented with the five image scales and horizontal flipping, namely ten augmented images for each image are passed in the network. We use the mean output of the model branch, including three refinement classification scores and the regression branch score, as the final score of each augmented image. Then the outputs are average. And the IoU threshold of the  Non-Maximum Suppression (NMS) is set to 0.3. Following \cite{WSOD2,C_WSL,E2E}, we use a regression branch to further improve the performance instead of the Fast R-CNN.

Our experiments are implemented based on PyTorch deep learning framework. And our experiments are running on an NVIDIA RTX 2080Ti GPU.

\begin{figure}[hb]
	\begin{center}
		\label{fig3a}\includegraphics[width=0.23\textwidth]{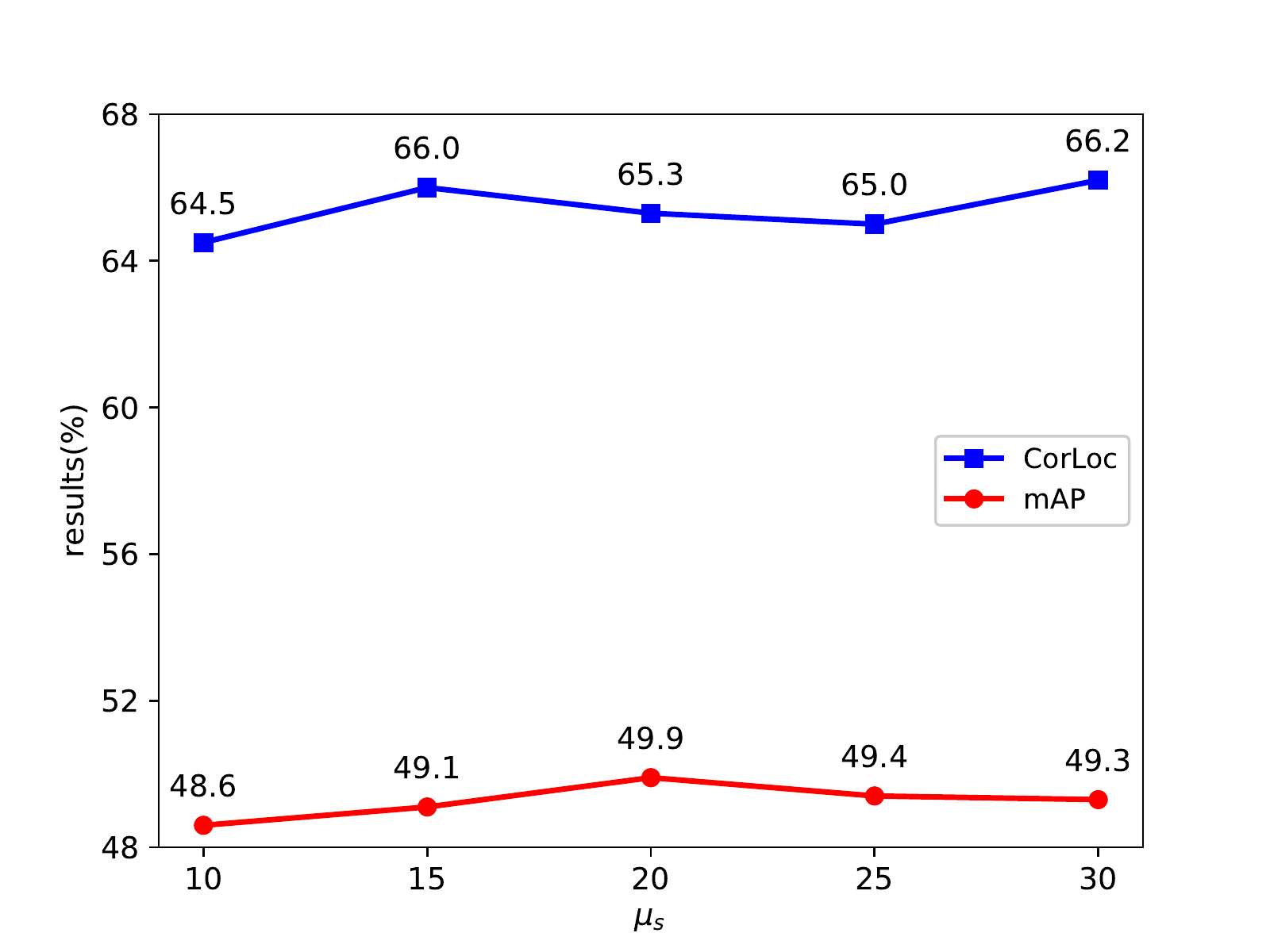}
		\label{fig3b}\includegraphics[width=0.23\textwidth]{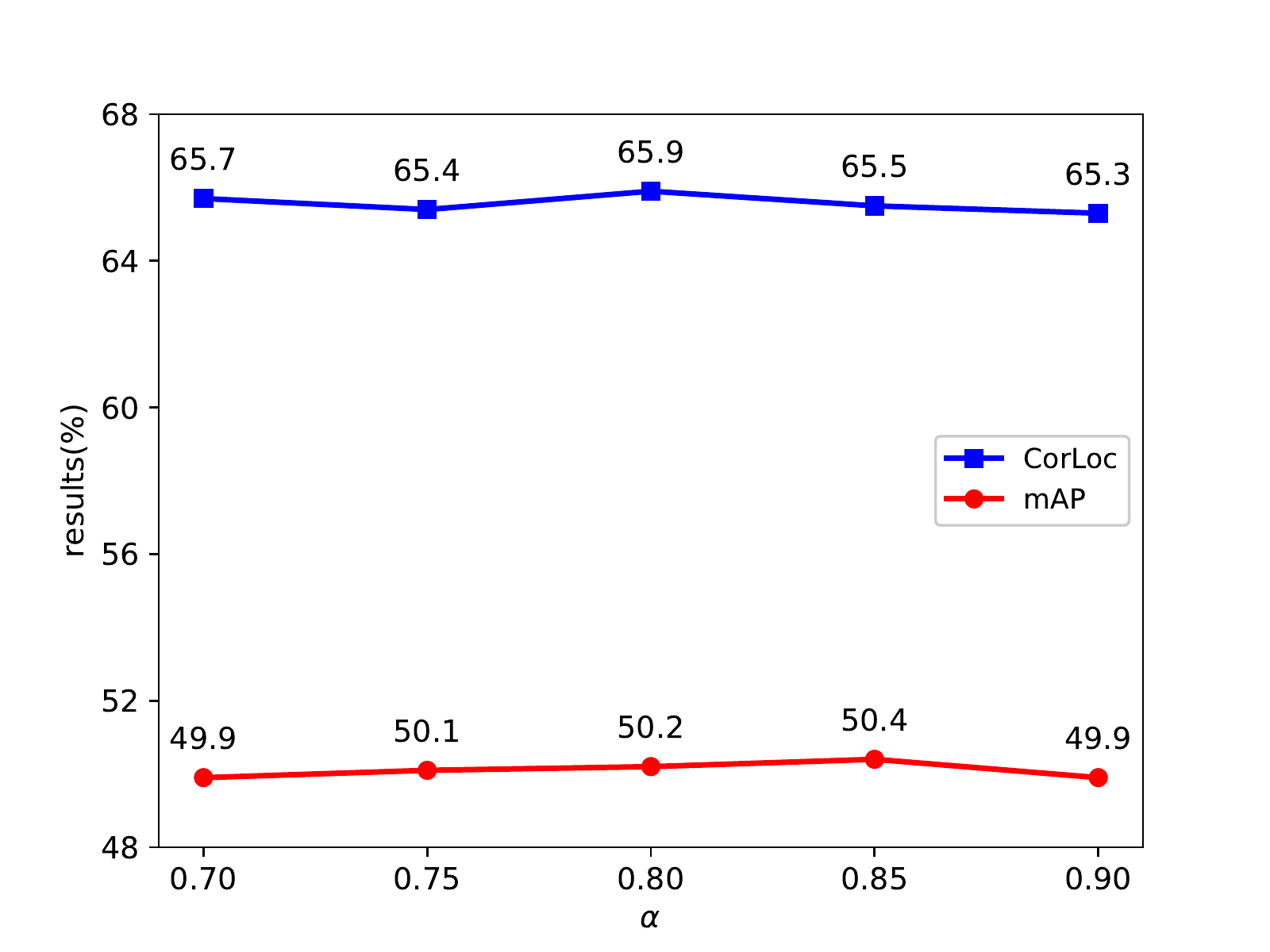}
	\end{center}
	\caption{ Detection performance on the PASCAL VOC 2007 $test$ set for using different values of hyper-parameter  $\mu_s$ and hyper-parameter $\alpha$ in progressive instance balance. }
	\label{fig3}
\end{figure}

\begin{table}[hb]
	\caption{Ablation study of different sampling methods in the PIB module. NPI denotes the neglect of positive instances.}
	\label{table4}
	\begin{center}
		\begin{tabular}{cc}
			\toprule
			Method                            & mAP \\ \hline
			Random sampling                   &   48.2        \\
			IoU-balanced sampling            &  48.6        \\
			Progressive mixed sampling       & 49.7             \\
			Random sampling+NPI                   & 49.5  \\
			IoU-balanced sampling+NPI  &  49.0   \\
			Progressive mixed sampling+NPI (ours)  &  $\textbf{50.4}$  \\
			\bottomrule 
		\end{tabular}
		
	\end{center}
\end{table}

\subsection{Ablation experiments}

We perform ablation experiments on PASCAL VOC 2007 to evaluate the contributions of the modules introduced above. Varying from the original OICR, our baseline ignores the negative instances with the IoU less than $\lambda_{ig}=0.1$ \cite{RegionProposal}.

\subsubsection{The influence of progressive instance balance}

The influence of the sampling ratio $\mu_s$ in the PIB module is analyzed. From Fig. \ref{fig3a}, it can be seen that setting $\mu_s$ to $20$ obtains the best performance. And if $\mu_s$ is too small, the detection performance will be down. It verifies that the distribution of negative instances is messy at the beginning of reselecting and the model can not accurately sample appropriate negative instances. Based on this result, $\mu_s$ is set to 20 for the other experiments. In addition, different $\alpha$ values in PIB using progressive mixed sampling are shown in Fig. \ref{fig3b}. When $\alpha$ is equal to 0.85, the model achieves the best performance.

As shown in Table \ref{table3} and \ref{table4}, we test different sampling methods to validate the effectiveness of the PIB module. Compared to the baseline, using random sampling and IoU-balanced sampling without the neglect of positive instances (NPI) could bring a $0.3\%$ and $0.7\%$ performance gain respectively. With NPI, both random sampling and IoU-balanced sampling could further boost the performance to $49.5\%$ and $49.0\%$ mAP respectively, which confirms our theory in Section 3.4.1. Furthermore, compared to these two methods, ``progressive mixed sampling+NPI'' could achieve much better performance, which is $2.2\%$ superior to the baseline.

\begin{figure}[hb]
	\begin{center}
		\label{fig4a}\includegraphics[width=0.23\textwidth]{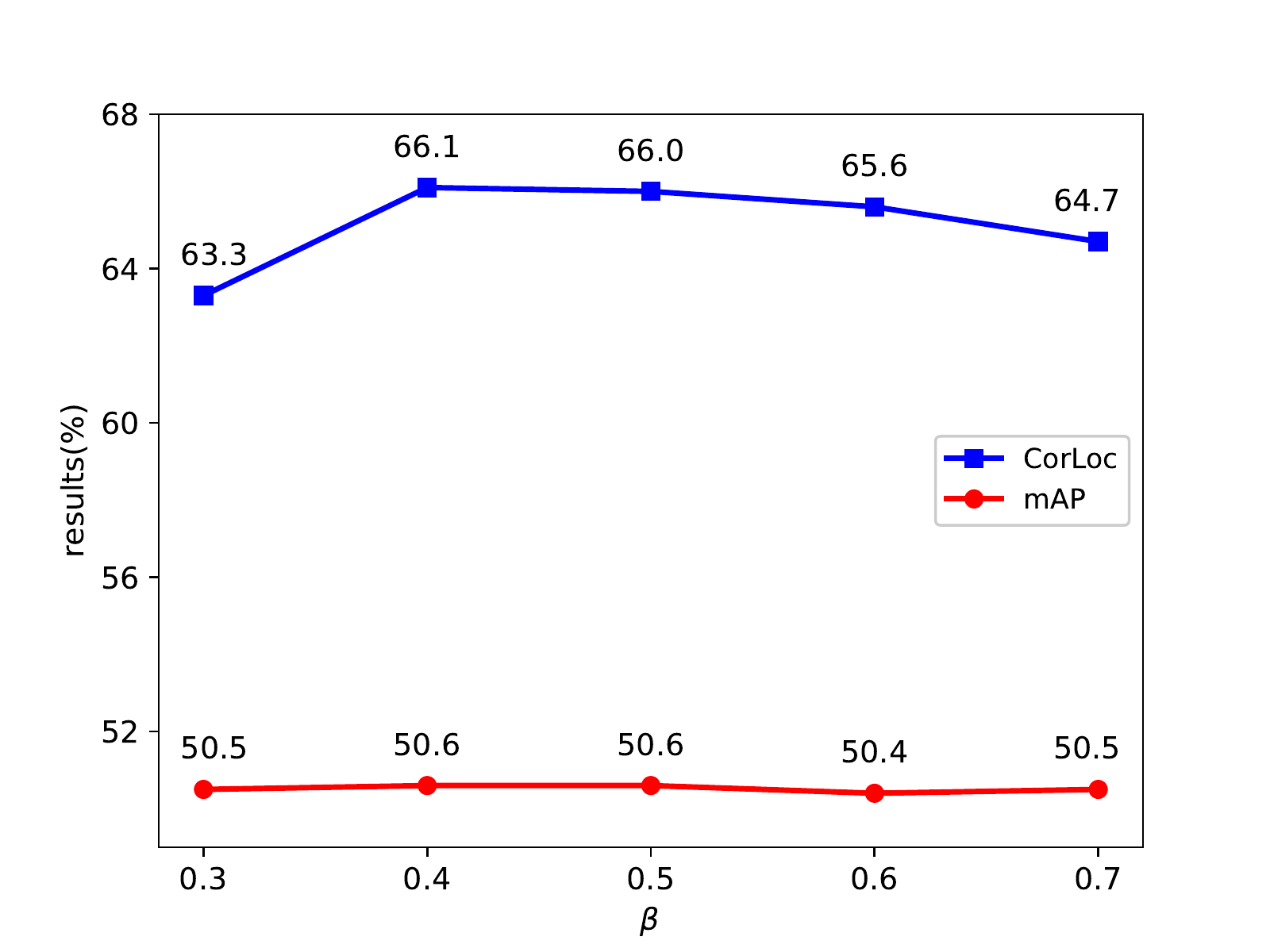}
		\label{fig4b}\includegraphics[width=0.23\textwidth]{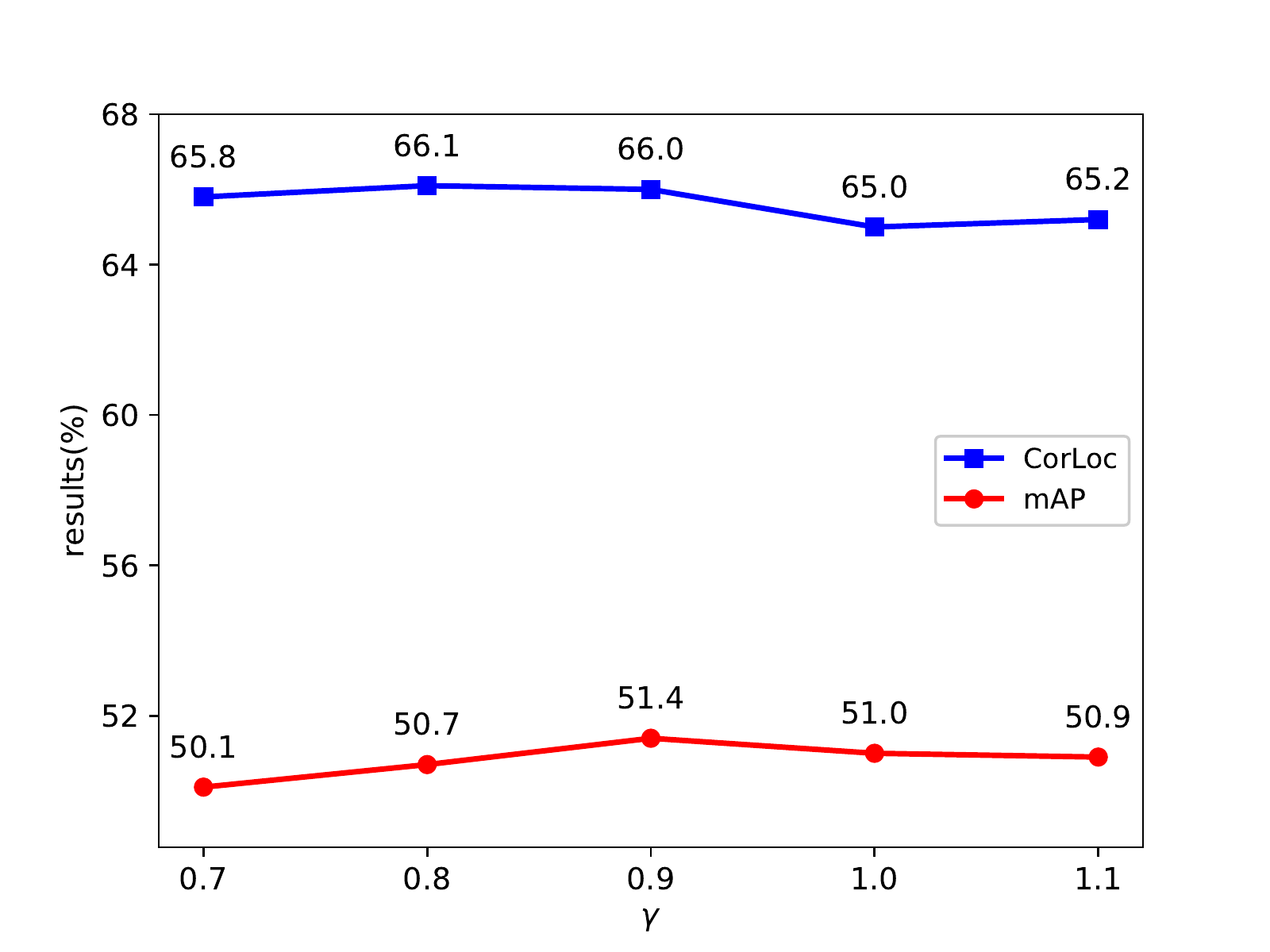}
	\end{center}
	\caption{ Detection performance on the PASCAL VOC 2007 $test$ set for using different values of hyper-parameter  $\beta$ and hyper-parameter $\gamma$ in progressive instance reweighting. }
	\label{fig4}
\end{figure}

\subsubsection{The influence of progressive instance reweighting}
As shown in Fig. \ref{fig4a}, the influence of PIR with different $\beta$ in Eq. (\ref{eq12}) is analyzed on the basis of the baseline during the whole training, which means Eq. (\ref{eq13}) is not used. Each value could improve the baseline to better performance. Both $\beta=0.4$ and $\beta=0.5$ boost the baseline to $50.6\%$ mAP and their CorLoc are $66.1\%$ and $66.0\%$ respectively. It confirms that the weight of positive instances is small during the normal training phase in the baseline and it needs to be reweighted. However, from Table \ref{table3}, combining PIB with PIR w/o AT , ``+PIB+PIR w/o AT'', only achieves $48.4\%$ mAP. And ``+PIB+PIR'' can further boost the performance to $51.4\%$ mAP, which is $0.8\%$ better than the ``+PIR w/o AT''. It verifies that the weight of positive instances should be reduced while the network uses the PIB module with the number of negative instances decreased. 
In addition,  the influence of the parameter $\gamma$ in Eq. (\ref{eq13}) is also studied as shown in Fig. \ref{fig4b}. It can be seen that setting $\gamma=0.9$ could lead to the best performance and the model is not too sensitive to $\gamma$.

\begin{table*}[htp]
	\centering
	\caption{Detection average precision ($\%$) for different methods on the PASCAL VOC 2012 test set.}
	\label{table6}
	\resizebox{\textwidth}{25mm}{
		\begin{tabular}{lccccccccccccccccccccc}
			\toprule
			Method & aero  & bike  & bird  & boat  & bottle & bus   & car   & cat   & chair & cow   & table & dog   & horse & mbike & person & plant & sheep & sofa  & train & tv    & mAP \\
			\midrule
			OICR \cite{OICR} & 67.7  & 61.2  & 41.5  & 25.6  & 22.2  & 54.6  & 49.7  & 25.4  & 19.9  & 47.0  & 18.1  & 26.0  & 38.9  & 67.7  & 2.0   & 22.6  & 41.1  & 34.3  & 37.9  & 55.3  & 37.9  \\
			WS-JDS \cite{WS-JDS} & -     & -     & -     & -     & -     & -     & -     & -     & -     & -     & -     & -     & -     & -     & -     & -     & -     & -     & -     & -     & 39.1  \\
			TS2C  \cite{TS2C}  & 67.4  & 57.0  & 37.7  & 23.7  & 15.2  & 56.9  & 49.1  & \textbf{64.8}  & 15.1  & 39.4  & 19.3  & 48.4  & 44.5  & 67.2  & 2.1   & 23.3  & 35.1  & 40.2  & 46.6  & 45.8  & 40.0  \\
			PCL \cite{PCL}  & 58.2  & 66.0  & 41.8  & 24.8  & 27.2  & 55.7  & 55.2  & 28.5  & 16.6  & 51.0  & 17.5  & 28.6  & 49.7  & 70.5  & 7.1   & 25.7  & 47.5  & 36.6  & 44.1  & 59.2  & 40.6  \\
			MELM \cite{MELM}  & -     & -     & -     & -     & -     & -     & -     & -     & -     & -     & -     & -     & -     & -     & -     & -     & -     & -     & -     & -     & 42.4  \\
			C-WSL \cite{C_WSL} & 74.0  & 67.3  & 45.6  & 29.2  & 26.8  & \textbf{62.5}  & 54.8  & 21.5  & \textbf{22.6}  & 50.6  & 24.7  & 25.6  & 57.4  & 71.0  & 2.4   & 22.8  & 44.5  & 44.2  & 45.2  & \textbf{66.9}  & 43.0  \\
			OAIL  \cite{OAIL} & 70.2  & 61.3  & 43.8  & 28.9  & 23.5  & 54.0  & 52.1  & 55.2  & 19.1  & 51.0  & 15.6  & 52.6  & 56.6  & 68.9  & \textbf{22.0}  & 21.7  & 43.6  & 37.0  & 34.8  & 56.3  & 43.4  \\
			DPS \cite{DPS}  & -     & -     & -     & -     & -     & -     & -     & -     & -     & -     & -     & -     & -     & -     & -     & -     & -     & -     & -     & -     & 43.8  \\
			MIL-PCL+GAM \cite{E2E} & 60.4  & 68.6  & 51.4  & 22.0  & 25.9  & 49.4  & 58.4  & 62.1  & 14.5  & 58.8  & 24.6  & \textbf{60.4}  & \textbf{64.3}  & 70.3  & 9.4   & 26.0  & 47.7  & \textbf{45.5}  & 36.7  & 55.8  & 45.6  \\
			OPG \cite{OPG}   & -     & -     & -     & -     & -     & -     & -     & -     & -     & -     & -     & -     & -     & -     & -     & -     & -     & -     & -     & -     & 46.2  \\
			OIM  \cite{OIM}  & -     & -     & -     & -     & -     & -     & -     & -     & -     & -     & -     & -     & -     & -     & -     & -     & -     & -     & -     & -     & 46.4  \\
			Boosted\_OICR \cite{BoostOICR} & 73.1  & 67.7  & \textbf{51.8}  & 29.8  & 31.8  & 60.0  & 59.7  & 33.2  & 18.9  & 60.7  & 22.7  & 46.8  & 60.2  & 73.9  & 3.9   & 26.6  & 51.7  & 40.7  & 60.2  & 61.3  & 46.7  \\
			C-MIL \cite{C-MIL} & -     & -     & -     & -     & -     & -     & -     & -     & -     & -     & -     & -     & -     & -     & -     & -     & -     & -     & -     & -     & 46.7  \\
			PredNet \cite{Pred_Net} & -     & -     & -     & -     & -     & -     & -     & -     & -     & -     & -     & -     & -     & -     & -     & -     & -     & -     & -     & -     & 48.4  \\
			Baseline & 73.1  & 67.0  & 42.8  & \textbf{32.0}  & 30.2  & 61.3  & 58.2  & 18.7  & 18.5  & 57.5  & 20.5  & 34.0  & 51.4  & 72.5  & 5.8   & 27.9  & 51.0  & 31.1  & 58.5  & 59.4  & 43.6  \\
			OPIS (ours) & 73.3  & 71.0  & 50.9  & 31.1  & 31.3  & 59.7  & 60.0  & 18.5  & 20.7  & 59.2  & \textbf{25.1}  & 32.0  & 52.2  & 74.5  & 4.0   & \textbf{29.3}  & 53.1  & 38.3  & 54.9  & 59.6  & 44.9  \\
			OPIS+REG (ours) & \textbf{74.5}  & \textbf{72.0}  & 50.2  & 29.2  & \textbf{36.9}  & 61.1  & \textbf{63.2}  & 23.8  & 22.0  & \textbf{64.2}  & 24.6  & 55.9  & 50.4  & \textbf{74.7}  & 17.5  & \textbf{29.3}  & \textbf{54.9}  & 38.4  & \textbf{63.1}  & 63.1  & \textbf{48.5}  \\
			\bottomrule
		\end{tabular}
	}
\end{table*}

\begin{table*}[htp]
	\centering
	\caption{Detection CorLoc ($\%$) for different methods on the PASCAL VOC 2012 trainval set.}
	\label{table7}
	\resizebox{\textwidth}{20mm}{
		\begin{tabular}{lccccccccccccccccccccc}
			\toprule
			Method & aero  & bike  & bird  & boat  & bottle & bus   & car   & cat   & chair & cow   & table & dog   & horse & mbike & person & plant & sheep & sofa  & train & tv    & CorLoc \\
			\midrule
			OICR \cite{OICR} & 86.2  & 84.2  & 68.7  & 55.4  & 46.5  & 82.8  & 74.9  & 32.2  & 46.7  & 82.8  & 42.9  & 41.0  & 68.1  & 89.6  & 9.2   & 53.9  & 81.0  & 52.9  & 59.5  & 83.2  & 62.1  \\
			WS-JDS \cite{WS-JDS} & -     & -     & -     & -     & -     & -     & -     & -     & -     & -     & -     & -     & -     & -     & -     & -     & -     & -     & -     & -     & 63.5  \\
			TS2C \cite{TS2C} & 79.1  & 83.9  & 64.6  & 50.6  & 37.8  & 87.4  & 74.0  & \textbf{74.1}  & 40.4  & 80.6  & 42.6  & 53.6  & 66.5  & 88.8  & 18.8  & 54.9  & 80.4  & 60.4  & 70.7  & 79.3  & 64.4  \\
			PCL \cite{PCL} & 77.2  & 83.0  & 62.1  & 55.0  & 49.3  & 83.0  & 75.8  & 37.7  & 43.2  & 81.6  & 46.8  & 42.9  & 73.3  & 90.3  & 21.4  & 56.7  & 84.4  & 55.0  & 62.9  & 82.5  & 63.2  \\
			OAIL \cite{OAIL} & 86.5  & 82.1  & 67.2  & 58.7  & 48.9  & 80.5  & 75.6  & 62.3  & 46.0  & 81.9  & 40.0  & 64.2  & 82.4  & 88.2  & \textbf{44.2}  & 53.5  & 78.1  & 54.7  & 56.7  & 82.9  & 66.7  \\
			MIL-PCL+GAM \cite{E2E} & 80.2  & 83.0  & \textbf{73.1}  & 51.6  & 48.3  & 79.8  & 76.6  & 70.3  & 44.1  & 87.7  & 50.9  & \textbf{70.3}  & \textbf{84.7}  & 92.4  & 28.5  & 59.3  & 83.4  & \textbf{64.6}  & 63.8  & 81.2  & 68.7  \\
			OPG \cite{OPG} & -     & -     & -     & -     & -     & -     & -     & -     & -     & -     & -     & -     & -     & -     & -     & -     & -     & -     & -     & -     & 65.8  \\
			OIM \cite{OIM} & -     & -     & -     & -     & -     & -     & -     & -     & -     & -     & -     & -     & -     & -     & -     & -     & -     & -     & -     & -     & \textbf{69.5}  \\
			Boosted\_OICR \cite{BoostOICR} & -     & -     & -     & -     & -     & -     & -     & -     & -     & -     & -     & -     & -     & -     & -     & -     & -     & -     & -     & -     & 66.3  \\
			C-MIL \cite{C-MIL} & -     & -     & -     & -     & -     & -     & -     & -     & -     & -     & -     & -     & -     & -     & -     & -     & -     & -     & -     & -     & 67.4  \\
			PredNet \cite{Pred_Net} & -     & -     & -     & -     & -     & -     & -     & -     & -     & -     & -     & -     & -     & -     & -     & -     & -     & -     & -     & -     & \textbf{69.5}  \\
			Baseline & 87.3  & 84.8  & 65.5  & 58.1  & 53.6  & \textbf{88.8}  & 75.2  & 30.0  & 45.8  & 82.2  & 44.9  & 44.3  & 75.4  & 88.2  & 16.4  & 60.4  & 83.1  & 44.3  & 74.5  & 83.2  & 64.3  \\
			OPIS (ours) & 87.7  & 84.6  & 69.6  & \textbf{59.7}  & 53.8  & 87.7  & 77.1  & 29.3  & 49.1  & 79.9  & 49.1  & 43.8  & 72.0  & 91.4  & 13.8  & 61.4  & 85.0  & 53.2  & 72.2  & 83.5  & 65.2  \\
			OPIS+REG (ours) & \textbf{88.5}  & \textbf{85.5}  & 71.2  & 59.1  & \textbf{58.3}  & 87.2  & \textbf{81.7}  & 33.6  & \textbf{50.4}  & \textbf{89.3}  & \textbf{51.0}  & 63.8  & 69.4  & \textbf{92.7}  & 35.9  & \textbf{62.1}  & \textbf{85.9}  & 52.4  & \textbf{80.2}  & \textbf{85.0}  & 69.2  \\
			\bottomrule
		\end{tabular}
	}
\end{table*}

\begin{figure*}[htp]
	\centering
	\includegraphics[width=7in]{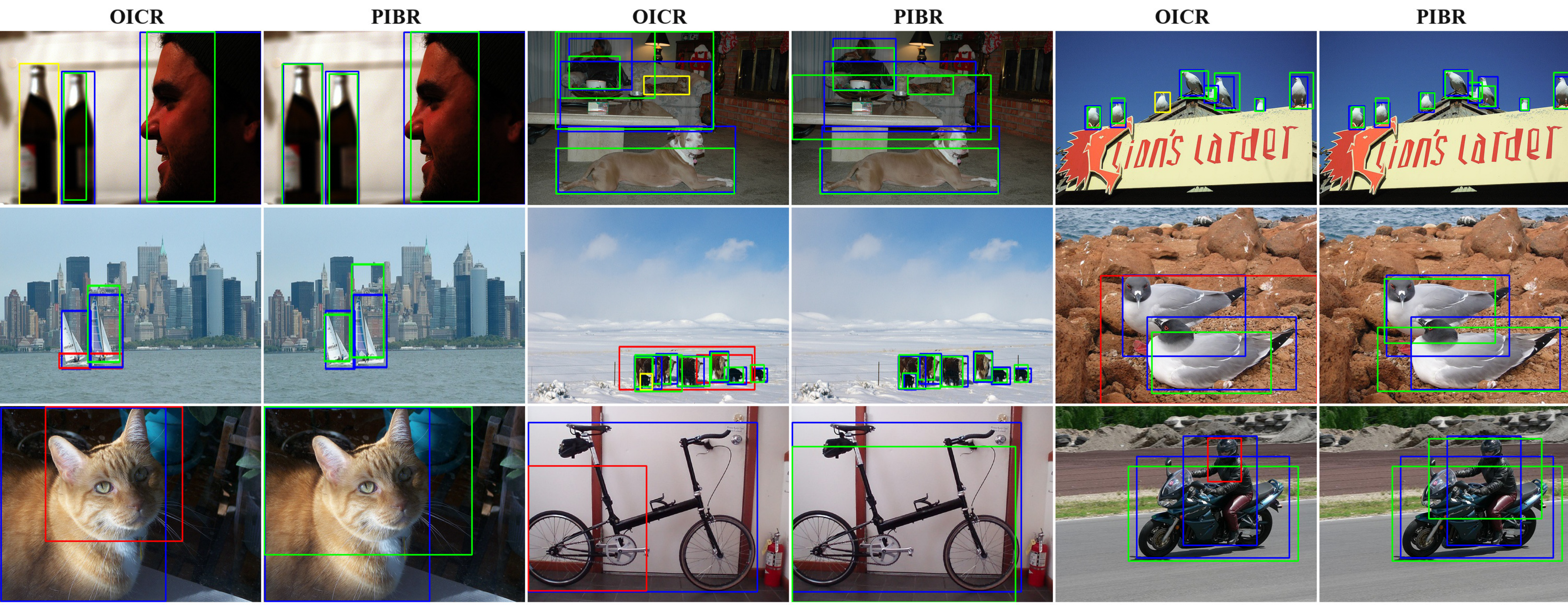}
	\caption{Some visualization comparisons between OICR and OPIS on the PASCAL VOC 2007 $test$ set. Green rectangles represent the successful detections (IoU$\geq$0.5), red rectangles represent the failed detections (IoU$<$0.5),  blue rectangles are the ground-truth with detection results, and yellow rectangles indicate the ground truths with no detection results.}
	\label{fig5}
\end{figure*}

\begin{figure*}[ht]
	\centering
	\includegraphics[width=7in]{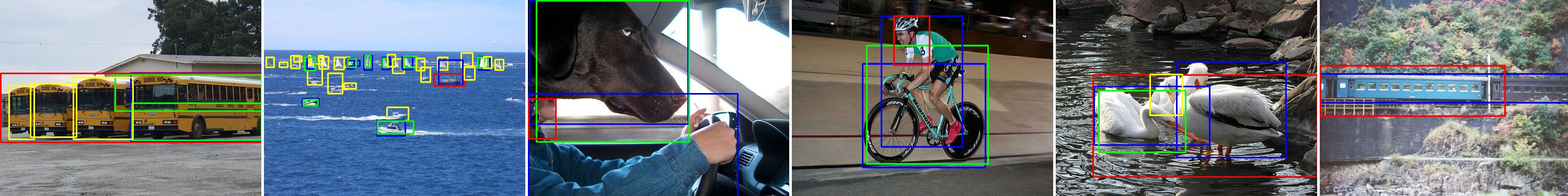}
	\caption{Some failed detection examples on the PASCAL VOC 2007 $test$ set. Green rectangles represent the successful detections (IoU$\geq$0.5), red rectangles represent the failed detections (IoU$<$0.5),  blue rectangles are the ground-truth with detection results, and yellow rectangles indicate the ground truths with no detection results.}
	\label{fig6}
\end{figure*}

\subsection{Runtime}

The runtime excluding proposal generation time and pretreatment time is shown in Table \ref{table5}. It can be seen that our method takes almost the same training time as the baseline because our method just involves resampling and reweighting in the supervision of instance classifier refinements and does not need an additional architecture. And it uses very little extra CPU usage. However, our method acquires much better detection performance than the baseline.

\begin{table}[h]
	\begin{center}
		\caption{Runtime comparisons between the baseline and the proposed method.}
		\label{table5}
		\begin{tabular}{cc}
			\toprule
			Method	& Training(sec/epoch)  \\
			\midrule
			Baseline &        1250.7                 \\
			OPIS (ours)     &        1254.1                 \\
			\bottomrule            
		\end{tabular}
	\end{center}
\end{table}

\subsection{Comparison with other methods}

The detection performance of OPIS and other methods for each class on PASCAL VOC 2007 is reported in Table \ref{table1}  and Table \ref{table2}. For fair comparisons, the methods that use  extra supervision beyond image-level labels are not listed. Noteworthily, our model significantly outperforms the baseline by $3.5\%$ on mAP and it could achieve $53.2\%$ mAP and $68.4\%$ CorLoc with the regression branch.  Compared with the other methods listed, our method outperforms these competitors. Our method does not need to add extra architecture and obtains much better results than the methods \cite{BoostOICR,OPG,OIM} based on the OICR network, e.g. Boosted OICR \cite{BoostOICR} adds a knowledge distillation module into the network, which is more complex than our model. Further, our method could outperform some methods using an ensemble of models. Even though our method is simple, the model achieves promising results.

We further perform experiments on PASCAL VOC2012 and the results are shown in Table \ref{table6} and Table \ref{table7}. Obviously, OPIS could achieve  $44.9\%$ mAP and $65.2\%$ CorLoc and it could reach $48.5\%$ mAP and $69.2\%$ CorLoc with the regression branch. And it also demonstrates the effectiveness of the proposed method.

\subsection{Qualitative results}

Qualitatively, some visualization comparisons between OICR (baseline) and the proposed OPIS method on the PASCAL VOC 2007 $test$ set are shown in Fig. (\ref{fig5}). The OPIS method ameliorates the situations summarized in \cite{Wetectron} of failed detections in WSOD. In the first row, it can be seen that some objects, missed by OICR, could be detected by OPIS. In the second row, OICR would gather many spatially adjacent instances belonging to the same class in one big bounding box. However, OPIS could differentiate each object better. In the last row, OICR would select the instance located at the most discriminative parts of objects as the detection results. OPIS could select boxes covering the whole objects, which has a more comprehensive detection of objects for some categories.

Then some failed detection examples are shown in Fig. (\ref{fig6}). It can be observed that the problem of part domination and instance ambiguity is not solved completely though the proposed method alleviates it. For non-rigid objects, part domination still exists, e.g. the model detects a face as a person. Dense groups of instances are missed or predicted as a big bounding box, especially small boats. Therefore, it still remains a challenge to solve these problems. And we consider that positive instances are still inclined to focus on the most discriminative region of the object and miss some true positive instances with low scores because of the top-scoring principle. Therefore, we will pay attention to exploring other cues to select more accurate positive instances.

\section{Conclusion}
In this paper, we propose an effective strategy, online progressive instance-balanced sampling, for weakly supervised object detection. First, progressive instance balance aims to balance positive instances and negative instances progressively and this method would also gradually guide the model to mine hard negative instances. Meanwhile, it would neglect certain positive instances when the instance cluster center has no negative instances. Second, positive instances are reweighted by the scores and IoUs of adjacent instance classifier refinement stages to improve the focus of the model on positive instances. Experiments demonstrate that the proposed approach, without adding any extra network architectures, has achieved an inspiring performance outperforming many existing state-of-the-art results on the PASCAL VOC 2007 and 2012 benchmarks. Compared to the baseline, the appendant training overheads of the proposed method are very small. In the future,  we will pay attention to exploring other cues to select more accurate positive instances to further improve the detection performance and studying the generalization ability of OPIS for other weakly supervised tasks.

\section*{Acknowledgments}
This work was supported by the National Key Research and Development Program of China (Grant 2018YFB1307400) and the National Natural Science Foundation of China (Grant 61873267).

\bibliographystyle{IEEEtran}
\bibliography{mybibfile}

\end{document}